\documentclass[letterpaper]{article}
\usepackage{aaai2027}
\usepackage[hyphens]{url}
\usepackage{graphicx}
\usepackage{natbib}
\usepackage{caption}
\usepackage{booktabs}
\usepackage{amsmath}
\usepackage{amssymb}

\title{RiskWorld: Object-Centric Latent World Modeling for Autonomous Driving Risk Identification}

\author{
Jingzheng Li\textsuperscript{\rm 1},
Yufei~Ge\textsuperscript{\rm 2},
Qianren~Mao, \textsuperscript{\rm 1},
Zhijun Chen\textsuperscript{\rm 3},
Bing Li\textsuperscript{\rm 4},
Xingyu Peng\textsuperscript{\rm 1,\rm 5},\\
Baochang~Zhang\textsuperscript{\rm 1,\rm 5},
Xianglong Liu\textsuperscript{\rm 1,\rm 5}
}

\affiliations{
\textsuperscript{\rm 1}Zhongguancun Laboratory, Beijing, China\\
\textsuperscript{\rm 2}Tianjin University, Beijing, China\\
\textsuperscript{\rm 3}Hong Kong Polytechnic University, Hong Kong, China\\
\textsuperscript{\rm 4}Nanyang Technological University, Singapore\\ 
\textsuperscript{\rm 5}Beihang University, Beijing, China
}

\begin{document}

\maketitle

\begin{abstract}
Autonomous driving risk identification aims to determine which observed object is likely to become safety-critical to the ego vehicle. Existing approaches typically predict scene-level accidents, infer risk objects indirectly from ego behavior, or apply geometric checks after trajectory forecasting, without directly using predicted ego--object relations for risk-source localization. We propose RiskWorld, an object-centric latent world model that identifies risk from the imagined evolution of each candidate relative to the ego vehicle. RiskWorld combines pretrained predictive video representations with structured ego--object histories, contextualizes observed interactions, and rolls relation-aware object states into the future using RSSM-style latent dynamics. It decodes the rollout into object-level risk scores, supported by auxiliary future-relation and temporal-risk predictions. Inference uses only observations up to the current time, while logged futures provide training supervision. On RiskBench, RiskWorld achieves the best overall F1 of 63.0\% and the lowest false-alarm rate of 2.1\%. Further analyses show that the learned rollout captures the evolution of object-level risk before critical events, while RiskWorld's selections preserve planning-critical information under filtered observation.

\end{abstract}

\section{Introduction}
Safe autonomous driving requires a risk monitor that can answer a question beyond generic scene understanding: among surrounding traffic participants and obstacles, which object is likely to become safety-critical to the ego vehicle? This decision is needed before a collision, near miss, or evasive maneuver becomes visually obvious. For example, a static obstacle may be harmless until the ego path converges with it, whereas a pedestrian that appears safe in the current frame may become critical under future motion. Driving risk is therefore object-specific and temporally evolving, and reliable localization should be grounded in how each candidate's relation to the ego vehicle is expected to develop.

\begin{figure}[t]
\centering
\includegraphics[width=0.45\textwidth]{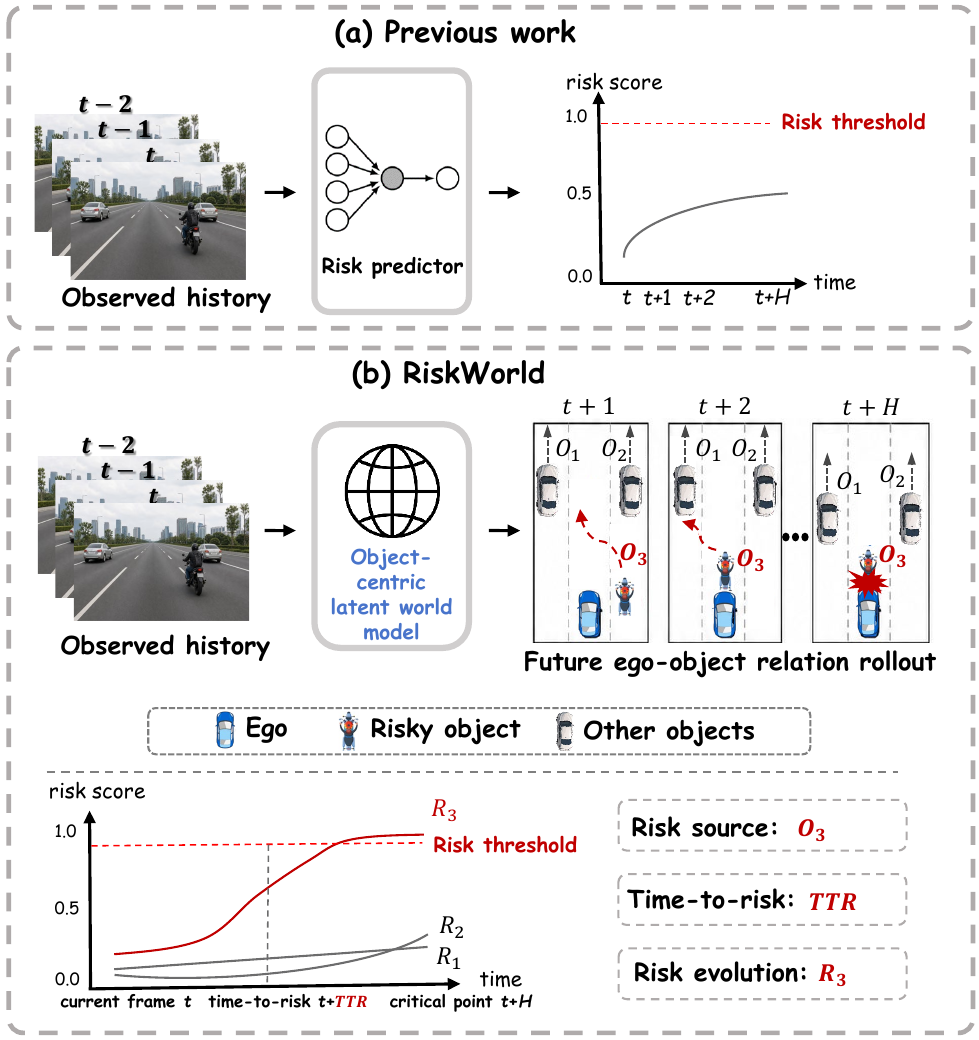}
\caption{Existing approaches predict scene-level risk, infer risk objects from ego behavior, or apply post-hoc geometric checks. RiskWorld instead predicts object-conditioned future ego--object relations and uses them to identify the risk source.}
\label{fig:intro}
\end{figure}

Existing methods mainly follow three paradigms. Accident anticipation methods estimate scene-level crash likelihood from dashcam videos \citep{crash2024,badas2025}; recent advances further predict future accident occurrence \citep{top2025}, forecast future latent scene representations \citep{flara2026}, or regularize the temporal evolution of risk scores \citep{riskprop2026}. Behavior-based methods infer risky objects indirectly from ego reactions such as braking or stopping \citep{bp2020,bcp2020}. Trajectory-geometric methods predict future motion and then apply distance, overlap, or collision rules to select a risk source \citep{mantra2020,qcnet2023}. These paradigms provide complementary signals, but future reasoning remains weakly coupled to object-level risk localization: accident predictors are largely scene-level, behavior-based localization is indirect, and trajectory-based risk is usually decided after forecasting. In particular, object-conditioned future ego--object relations are not directly optimized as predictive evidence for risk-source identification. Figure~\ref{fig:intro} contrasts these paradigms with our formulation.

We take a different view: risk is not a static visual attribute but the outcome of an evolving ego--object relation. An object becomes safety-critical when its relative motion, distance, path conflict, or interaction with the ego vehicle trends toward an unsafe state. A risk monitor should therefore predict object-conditioned future relations before assigning object-level risk.

World models and predictive representations provide a natural mechanism for this formulation. Recent driving world models roll compact scene states forward to forecast occupancy, point clouds, videos, or controllable scene evolution \citep{driveoccworld2025,vidar2024,driveworld2024,vista2024,drivedreamer2024,genad2024}. Their endpoints, however, are typically generation, simulation, occupancy forecasting, representation learning, or planning. Object-level risk identification instead calls for task-oriented latent prediction: rolling compact object states forward under ego--object relation and risk supervision rather than reconstructing an entire future scene.

We propose RiskWorld, a task-oriented latent world model for object-level driving risk identification. RiskWorld adapts pretrained predictive video representations \citep{vjepa2024,vjepa22025}, combines them with structured ego--object histories, and constructs relation-aware object states. RSSM-style latent dynamics \citep{dreamerv32023} roll these states into the future, while relation and risk decoders produce object risk scores together with auxiliary future-relation and temporal-risk predictions. Inference uses only current and historical observations.

Our contributions are:
\begin{itemize}
\item We formulate object-level driving risk identification as future ego--object relation rollout, using predicted relation evolution as evidence for localizing the risk source.
\item We propose RiskWorld, an object-centric latent world model that combines pretrained predictive video representations, relation-aware scene encoding, RSSM-style latent dynamics, and risk-oriented supervision.
\item On RiskBench, RiskWorld improves overall F1 by 1.2 points over the strongest baseline and achieves the lowest FA of 2.1\%; additional analyses examine temporal separation and planning relevance.
\end{itemize}

\section{Related Work}
\subsection{Risk Identification and Accident Anticipation}

Accident anticipation predicts impending collisions from observed dashcam video. Early methods estimate scene-level risk with spatiotemporal or agent-centric attention \citep{dsa2016,rrl2017,crash2024}. BADAS adapts predictive video representations to ego-centric collision prediction, and BADAS-2.0 scales this framework with real-time object-centric explanations \citep{badas2025,badas22026}. Recent methods make the temporal target more explicit: TOP predicts accident scores at multiple future horizons, FLaRA classifies predicted future latent representations, and RiskProp propagates collision-anchored supervision to learn temporally structured risk scores \citep{top2025,flara2026,riskprop2026}. These works establish future-aware accident anticipation, but their primary prediction remains a scene-level collision score; object attention provides explanatory evidence rather than an object-risk target.

Object-level risk identification instead asks which participant constitutes the risk source. Behavior-based approaches such as BP, BCP, and DROID infer risky objects from predicted ego behavior, while RiskBench standardizes localization, anticipation, and planning-aware evaluation across scenario types \citep{bp2020,bcp2020,droid2023,riskbench2024}. Motion forecasting offers another route by predicting agent trajectories followed by distance or collision checks \citep{mantra2020,qcnet2023}; MC-Risk extends this route with planner-aligned analytic risk fields on the RiskBench collision subset \citep{mcrisk2026}. Conformal Risk Tube Prediction (CRTP) instead predicts calibrated object-level risk intervals under spatiotemporal uncertainty \citep{crtp2026}. RiskWorld addresses a complementary gap: it learns object-conditioned latent dynamics supervised by future ego-relative geometry and decodes risk-source scores from the resulting relation rollout.

\subsection{World Models and Predictive Representations}

World models learn predictive states that can be rolled forward. PlaNet introduced the recurrent state-space model (RSSM), which combines deterministic recurrent and stochastic latent states, and DreamerV3 extends this family to imagination-based control across domains \citep{planet2019,dreamerv32023}. In parallel, V-JEPA and V-JEPA2 learn predictive video representations by forecasting latent targets rather than reconstructing future pixels \citep{vjepa2024,vjepa22025}. RiskWorld draws on these complementary ideas: frozen predictive visual features encode the observed context, while RSSM-style dynamics roll object states forward.

Driving world models instantiate future prediction through occupancy and flow \citep{occworld2024,driveoccworld2025}, visual point clouds and 4D scene representations \citep{vidar2024,driveworld2024}, or controllable video generation \citep{drivedreamer2024,vista2024}. Latent models such as LAW and World4Drive predict future representations to support trajectory generation or selection \citep{law2025,world4drive2025}, whereas DriveLaW connects video generation and motion planning through a shared latent world \citep{drivelaw2026}. Their primary endpoints are scene forecasting, generation, simulation, or planning. RiskWorld instead specializes latent rollout for object-level risk-source identification, with future relation and temporal targets providing task-oriented supervision and evidence.

\begin{figure*}[t]
\centering
\includegraphics[width=0.98\textwidth]{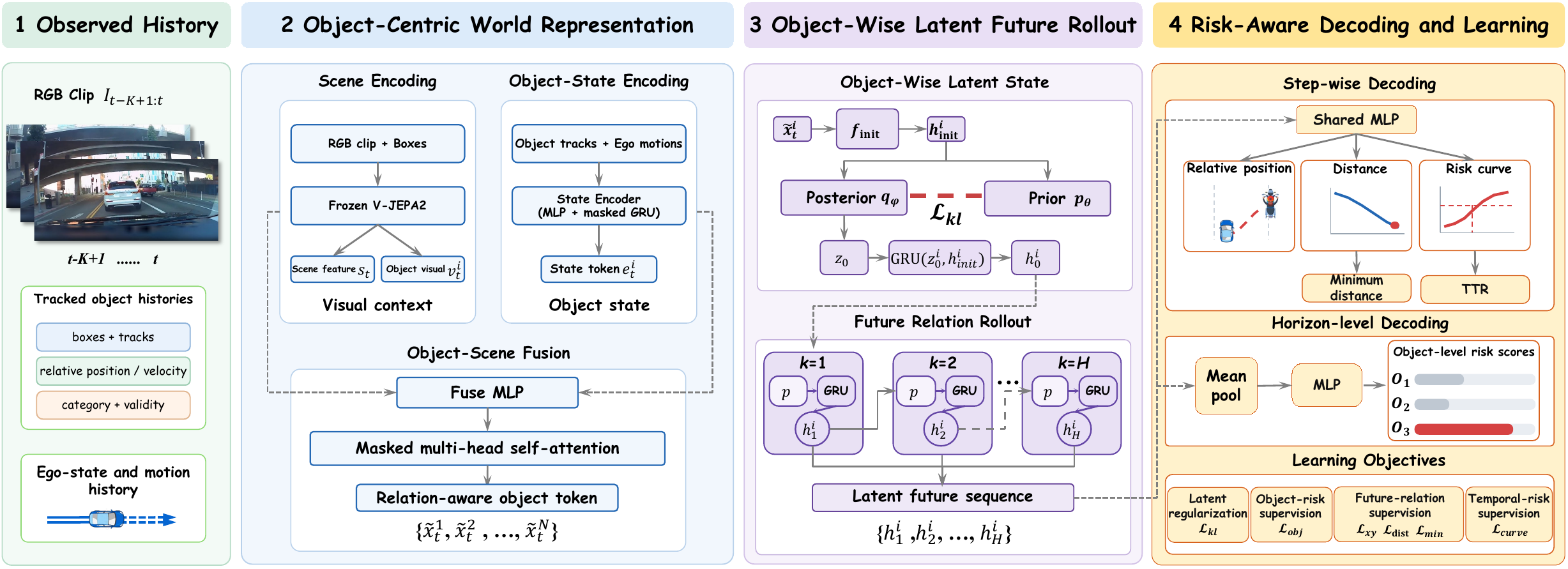}
\caption{Overview of RiskWorld. Observed RGB frames, object tracks, and ego motion are encoded into relation-aware object states, rolled forward with object-wise RSSM-style latent dynamics, and decoded into risk-source scores with auxiliary future-relation and temporal-risk predictions. Future annotations supervise training only; inference uses observed history.}
\label{fig:riskworld}
\end{figure*}

\section{Method: RiskWorld}

\subsection{Problem Formulation}

At frame $t$, RiskWorld observes a $K$-frame history $\mathcal{H}_t$ comprising front-view RGB, tracked object states, and ego motion. Let $\mathcal{O}_t=\{o_t^1,\ldots,o_t^N\}$ be the current candidates and $\nu_t^i\in\{0,1\}$ their validity indicators. For each valid candidate, $y_t^i$ indicates whether it is the ego-relevant risk object, and RiskWorld estimates
\begin{equation}
r_t^i=\Pr(y_t^i=1\mid \mathcal{H}_t,\mathcal{O}_t).
\end{equation}
Rather than judging risk from frame $t$ alone, the model obtains $r_t^i$ from an $H$-step latent rollout of future ego--object relations. Logged future annotations supervise this rollout during training but are never used as inference inputs.
\subsection{Overview}

Figure~\ref{fig:riskworld} summarizes four stages. A frozen V-JEPA2 encoder extracts predictive scene context, while a structured state encoder preserves explicit object and ego motion. Fusion and masked relation attention produce one contextualized token per candidate. An object-wise latent dynamics module, following an RSSM-style deterministic--stochastic design, then rolls each token into a latent future. Finally, step-wise heads decode future relation and risk trajectories, while the pooled future state produces the object risk score; minimum distance and time-to-risk are derived from the decoded trajectories.

\subsection{Object-Centric World Representation}

\textbf{Scene encoding.} A frozen V-JEPA2 encoder $\Phi_{\mathrm{w}}$ processes the observed clip $I_{t-K+1:t}$, with current boxes $B_t$ guiding object-level feature pooling. Averaging its spatiotemporal tokens gives a global scene feature $s_t$, while box pooling over the final temporal grid gives an object-aligned feature $v_t^i$:
\begin{equation}
(s_t,\{v_t^i\}_{i=1}^{N})=\Phi_{\mathrm{w}}(I_{t-K+1:t},B_t).
\end{equation}
These frozen features contribute appearance, layout, and motion context learned through predictive pretraining without reconstructing future pixels.

\textbf{Object-state encoding.} A multilayer perceptron (MLP) encodes each candidate's current ego-relative state, including position, motion, box geometry, and category, while a masked gated recurrent unit (GRU) summarizes its observed track history. Fusing the two gives a structured object feature $e_t^i$; the same encoder produces an ego-motion feature $e_t^{\mathrm{ego}}$.

\textbf{Object-scene relational fusion.} A shared fusion MLP projects and combines $(e_t^i,v_t^i,s_t)$ into an object token $x_t^i$, while $(e_t^{\mathrm{ego}},s_t)$ gives an ego token $x_t^{\mathrm{ego}}$. Masked multi-head self-attention, denoted $\operatorname{RelAttn}$, then models ego--object and object--object interactions:
\begin{equation}
\resizebox{0.98\columnwidth}{!}{$\displaystyle
\left[\tilde{x}_t^1,\ldots,\tilde{x}_t^N\right]
=\operatorname{RelAttn}\!\left(
\left[x_t^{\mathrm{ego}},x_t^1,\ldots,x_t^N\right],
\left[1,\nu_t^1,\ldots,\nu_t^N\right]
\right).$}
\end{equation}
The second argument is the validity mask, and the outputs $\{\tilde{x}_t^i\}_{i=1}^{N}$ are relation-aware object tokens used to initialize the latent rollout.

\subsection{Object-Wise Latent Future Rollout}

\textbf{Object-wise latent state.} Rather than generating future images, RiskWorld adopts an RSSM-style dynamics design to evolve each relation-aware object token in a compact latent space. A learned initializer first maps $\tilde{x}_t^i$ to a deterministic state:
\begin{equation}
h_{\mathrm{init}}^i=f_{\mathrm{init}}(\tilde{x}_t^i).
\end{equation}
The latent dynamics combines a deterministic recurrent state $h$ with a stochastic state $z$, and the GRU integrates $z$ into $h$ at each step. At the observed boundary, the posterior $q_\phi$ grounds $z_0^i$ in the object token, while the prior $p_\theta$ predicts it from $h_{\mathrm{init}}^i$ alone. The initial latent state is
\begin{equation}
\resizebox{0.7\columnwidth}{!}{$\displaystyle
z_0^i\sim q_\phi(z_0^i\mid h_{\mathrm{init}}^i,\tilde{x}_t^i),\;
h_0^i=\operatorname{GRU}(z_0^i,h_{\mathrm{init}}^i).$}
\end{equation}
Here $\phi$ and $\theta$ parameterize diagonal-Gaussian posterior and prior distributions whose means and scales are predicted from their conditioning states.

\textbf{Future relation rollout.} Starting from the observation-grounded state $h_0^i$, RiskWorld recursively predicts $H$ future latent states using only the learned prior:
\begin{equation}
\begin{aligned}
z_k^i&\sim p_\theta(z_k^i\mid h_{k-1}^i),\\
h_k^i&=\operatorname{GRU}(z_k^i,h_{k-1}^i),
\quad k=1,\ldots,H.
\end{aligned}
\end{equation}
The sequence $\{h_k^i\}_{k=1}^{H}$ represents the imagined evolution of object $i$ relative to the ego vehicle. No future observation is provided, and dynamics parameters are shared across candidates; observed interaction context is encoded in $\tilde{x}_t^i$.

\textbf{Latent regularization.} We regularize the observation-conditioned posterior toward the initial prior using
\begin{equation}
\resizebox{0.98\columnwidth}{!}{$\displaystyle
\mathcal{L}_{kl}=
\mathbb{E}_{i:\nu_t^i=1}\,\mathrm{KL}\!\left[
q_\phi(z_0^i\mid h_{\mathrm{init}}^i,\tilde{x}_t^i)
\,\|\,
p_\theta(z_0^i\mid h_{\mathrm{init}}^i)
\right].$}
\end{equation}
This term prevents an unconstrained posterior and stabilizes latent initialization. Since the current object token is observed at inference, we initialize with the posterior mean and subsequently roll forward using prior means; stochastic samples are used during training.

\subsection{Risk-Aware Decoding and Learning}

\textbf{Risk and relation decoding.} A shared step-wise head $f_{\mathrm{step}}$ maps every future state to ego-relative position $\hat{\mathbf{p}}_{t+k}^i$, distance $\hat d_{t+k}^i$, and temporal risk probability $\hat R_{t+k}^i$. A horizon-level head $f_{\mathrm{pool}}$ predicts the object risk score from the mean future state $\bar h_t^i=H^{-1}\sum_{k=1}^{H}h_k^i$:
\begin{equation}
\resizebox{0.6\columnwidth}{!}{$\displaystyle
\begin{aligned}
(\hat{\mathbf{p}}_{t+k}^i,\hat d_{t+k}^i,\hat R_{t+k}^i)
&=f_{\mathrm{step}}(h_k^i),\\
r_t^i&=f_{\mathrm{pool}}(\bar h_t^i).
\end{aligned}
$}
\end{equation}
The score $r_t^i$ identifies \emph{which} object is risky and is supervised by $y_t^i$ using class-balanced binary cross entropy $\mathcal{L}_{obj}$. The step-wise outputs describe \emph{when} risk emerges and how the underlying ego--object relation evolves; minimum distance and time-to-risk are derived from these trajectories rather than predicted separately.

\textbf{Future-relation supervision.} Logged ego and object tracks provide future ego-relative position and distance targets $\mathbf{p}_{t+k}^{i,*}$ and $d_{t+k}^{i,*}$. Over valid future steps, $\hat d_{\min,t}^i=\min_k\hat d_{t+k}^i$ and $d_{\min,t}^{i,*}=\min_k d_{t+k}^{i,*}$. Smooth-$L_1$ regression on position, distance, and the derived minimum distance gives $\mathcal{L}_{xy}$, $\mathcal{L}_{dist}$, and $\mathcal{L}_{min}$, respectively. These targets encourage the latent rollout to preserve the geometry most relevant to risk.

\textbf{Temporal-risk supervision.} Let $i^*$ be the annotated risk object and $[a,b]$ its event interval, with $a=b$ for a point event. We define
\begin{equation}
\resizebox{0.95\columnwidth}{!}{$\displaystyle
R_{t+k}^{i,*}=
\begin{cases}
\exp\!\left[-\dfrac{(a-(t+k))_+}{T}\right],
& i=i^*,\ a-H\leq t+k\leq b,\\
0, & \text{otherwise},
\end{cases}
$}
\end{equation}
where $(u)_+=\max(u,0)$ and $T$ controls temporal decay. The curve rises toward $a$, remains one over an annotated behavior interval, and is zero for other objects. Class-balanced binary cross entropy between $\hat R_{t+k}^i$ and $R_{t+k}^{i,*}$ defines $\mathcal{L}_{curve}$. Time-to-risk is obtained from the first crossing of the same predicted curve:
\begin{equation}
\widehat{\mathrm{TTR}}_t^i
=\Delta t\min\{k:\hat R_{t+k}^i\geq\gamma_{\mathrm{ttr}}\},
\end{equation}
where $\Delta t$ is the duration of one rollout step. No value is returned if the predicted curve does not cross $\gamma_{\mathrm{ttr}}$.

\textbf{Joint objective and inference.} Combining the losses introduced above gives
\begin{equation}
\begin{aligned}
\mathcal{L}
=&\lambda_{obj}\mathcal{L}_{obj}
+\lambda_{xy}\mathcal{L}_{xy}
+\lambda_{dist}\mathcal{L}_{dist}+\lambda_{min}\mathcal{L}_{min}\\
&
+\lambda_{curve}\mathcal{L}_{curve}
+\beta\mathcal{L}_{kl},
\end{aligned}
\end{equation}
where validity masks restrict each term to available objects and future steps, the $\lambda$ coefficients balance the supervised targets, and $\beta$ weights latent regularization. At inference, RiskWorld thresholds $r_t^i$ to identify risky objects; the relation trajectory, risk curve, minimum distance, and curve-derived time-to-risk provide object-specific future evidence.
\begin{table*}[t]
\centering
\caption{Risk localization and anticipation results. P, R, PIC, and FA denote precision, recall, Progressive Increasing Cost, and false-alarm rate. F1 is micro-averaged over all scenario types. $^{*}$ denotes our object-level adaptations of BADAS and FLaRA.}
\label{tab:riskbench_localization}
\resizebox{\textwidth}{!}{
\begin{tabular}{lccccccccccc}
\toprule
 & \multicolumn{3}{c}{Interactive} & \multicolumn{3}{c}{Collision} & \multicolumn{3}{c}{Obstacle} & Non-inter. & All \\
\cmidrule(lr){2-4}\cmidrule(lr){5-7}\cmidrule(lr){8-10}\cmidrule(lr){11-11}\cmidrule(lr){12-12}
Method & P(\%) $\uparrow$ & R(\%) $\uparrow$ & PIC $\downarrow$ & P(\%) $\uparrow$ & R(\%) $\uparrow$ & PIC $\downarrow$ & P(\%) $\uparrow$ & R(\%) $\uparrow$ & PIC $\downarrow$ & FA(\%) $\downarrow$ & F1 $\uparrow$ \\
\midrule
Random & 9.5 & 50.0 & 16.7 & 19.2 & 49.8 & 15.7 & 7.2 & 50.3 & 11.4 & 80.8 & 15.1 \\
Range (5m) & 38.7 & 10.7 & 20.3 & 85.6 & 27.7 & 4.1 & 45.0 & 18.3 & 10.2 & 3.3 & 30.0 \\
Range (10m) & 43.6 & 69.1 & 7.1 & 69.3 & 62.4 & 3.8 & 33.3 & 88.5 & 0.8 & 15.2 & 53.6 \\
\midrule
Kalman filter & 37.0 & 54.0 & 15.4 & 59.9 & 58.1 & 4.2 & 35.3 & 61.2 & 10.2 & 18.8 & 46.7 \\
Social-GAN & 39.4 & 57.7 & 14.1 & 60.8 & 45.4 & 7.8 & 36.9 & 69.3 & 4.0 & 16.3 & 46.0 \\
MANTRA & 36.3 & 62.1 & 13.3 & 58.3 & 45.4 & 7.1 & 35.7 & 71.8 & \textbf{2.5} & 16.5 & 45.4 \\
QCNet & 40.0 & 56.6 & 14.1 & 60.3 & 45.4 & 7.3 & 38.2 & 68.9 & 4.4 & 14.5 & 46.4 \\
DSA & 54.7 & 19.7 & 20.9 & 79.4 & 45.9 & \textbf{3.7} & 54.2 & 47.2 & 17.5 & 3.3 & 46.8 \\
RRL & 49.4 & 15.4 & 22.4 & 87.3 & \textbf{64.6} & 4.6 & 47.1 & 19.0 & 16.0 & 5.1 & 48.6 \\
BP & 36.4 & 16.3 & 25.6 & 74.4 & 9.8 & 25.1 & 22.4 & 3.0 & 27.9 & 4.7 & 16.6 \\
BCP & 48.7 & 29.2 & 22.8 & 76.8 & 20.7 & 20.8 & 32.1 & 35.2 & 20.5 & 4.7 & 33.4 \\
BADAS$^{*}$ & 65.6 & 57.1 & 13.5 & 89.3 & 36.0 & 9.3 & 51.0 & 64.5 & 28.6 & 2.8 & 48.8 \\
FLaRA$^{*}$ & 64.2 & \textbf{73.6} & 12.6 & 89.9 & 41.5 & 6.1 & 65.4 & 67.8 & 15.6 & 4.1 & 61.8 \\
\midrule
RiskWorld & \textbf{70.2} & 66.5 & \textbf{10.5} & \textbf{91.5} & 40.4 & 8.9 & \textbf{67.3} & \textbf{72.0} & 15.7 & \textbf{2.1} & \textbf{63.0} \\
\bottomrule
\end{tabular}
}
\end{table*}
\section{Experiments}

We organize the evaluation around four research questions. \textbf{RQ1:} How effectively does RiskWorld localize risk sources across diverse driving scenarios compared with existing approaches? \textbf{RQ2:} Does RiskWorld capture how object-level risk evolves over the future horizon and distinguish the risk source before a critical event? \textbf{RQ3:} How do predictive world features, relational reasoning, latent rollout, and future supervision contribute to performance? \textbf{RQ4:} Do the objects selected by RiskWorld preserve planning-critical information when all other objects are masked?

\subsection{Experimental Setup}

\textbf{Dataset and protocol.} We evaluate on RiskBench \citep{riskbench2024}, which contains 6,916 CARLA scenarios spanning interactive conflicts, collisions, static obstacles, and non-interactive driving. We use its map-based train/validation/test split and front-view setting. Each input comprises RGB history, object boxes and tracks, and ego motion. Logged future tracks and event annotations supervise training only; test-time prediction is strictly causal and evaluated at every frame.

\textbf{Baselines.} We compare random and fixed-range rules; trajectory methods using Kalman filtering, Social-GAN, MANTRA, and QCNet \citep{kalman1960,socialgan2018,mantra2020,qcnet2023}; and visual/behavior methods DSA, RRL, BP, and BCP \citep{dsa2016,rrl2017,bp2020,bcp2020}. TOP, BADAS, FLaRA, and RiskProp predict scene-level risk and are not directly comparable under the object-level protocol \citep{top2025,badas2025,flara2026,riskprop2026}. We therefore adapt BADAS and FLaRA: BADAS$^{*}$ adds an object-risk head to predictive video features, whereas FLaRA$^{*}$ pools future latent tokens within each candidate box for object-risk classification. Both are trained and evaluated under the same protocol as RiskWorld.

\textbf{Metrics.} We report precision, recall, and overall micro-F1 for risk-source localization. Progressive Increasing Cost (PIC) assigns increasing weight to frame-level localization errors as the critical frame approaches, while the false-alarm rate (FA) is measured on non-interactive scenes. Future-risk prediction is evaluated using the horizon-wise Brier score, where lower is better. Planning-aware evaluation additionally reports Influenced Ratio (IR), the relative change in the closest ego--risk distance between full and filtered observations, and Collision Rate (CR). 
% Lower values are better for PIC, FA, IR, and CR.

\textbf{Implementation details.} We freeze a V-JEPA2 ViT-L encoder and use a $K=16$ frame observation history and an $H=60$ step future horizon (3\,s at 20 FPS). RiskWorld considers at most 64 candidates and uses 256-dimensional deterministic states, 64-dimensional stochastic states, and four relation-attention heads. We train for 10 epochs with AdamW, a batch size of 64, and an initial learning rate of $10^{-4}$. We set $\lambda_{obj}=1$, $\lambda_{curve}=0.2$, $\lambda_{xy}=\lambda_{dist}=0.05$, $\lambda_{min}=0.02$, and $\beta=10^{-3}$; auxiliary future losses are linearly warmed up over the first two epochs.
% These settings are fixed across all experiments.
We select $\eta=0.70$ on the validation set to maximize overall micro-F1 and use it for all test categories.

\begin{figure*}[t]
\centering
\begin{minipage}[t]{0.32\textwidth}
\vspace{0pt}
\centering
\includegraphics[width=\linewidth]{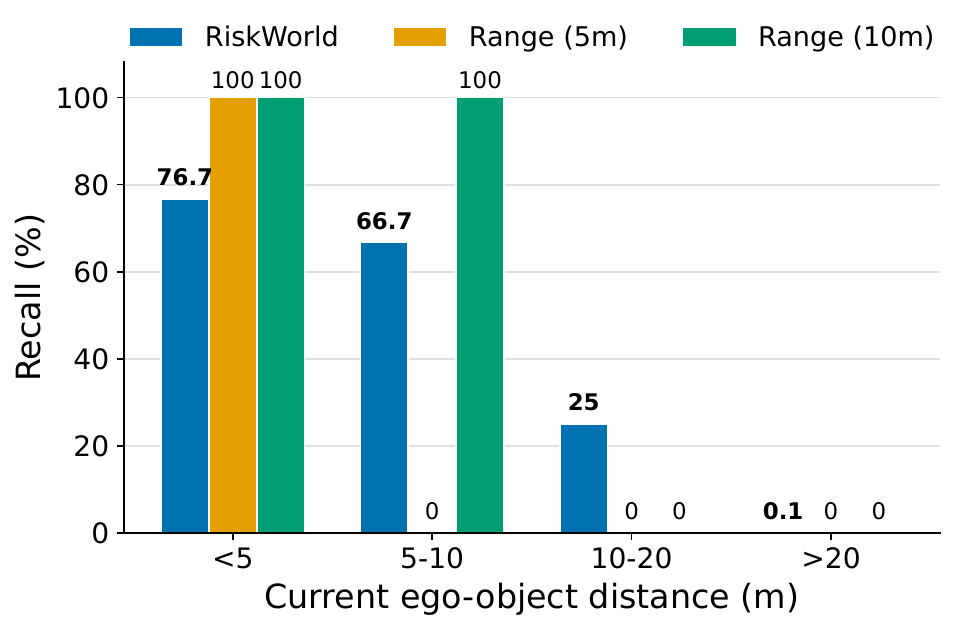}
\par\scriptsize (a) Recall on true risk objects
\end{minipage}\hfill
\begin{minipage}[t]{0.32\textwidth}
\vspace{0pt}
\centering
\includegraphics[width=\linewidth]{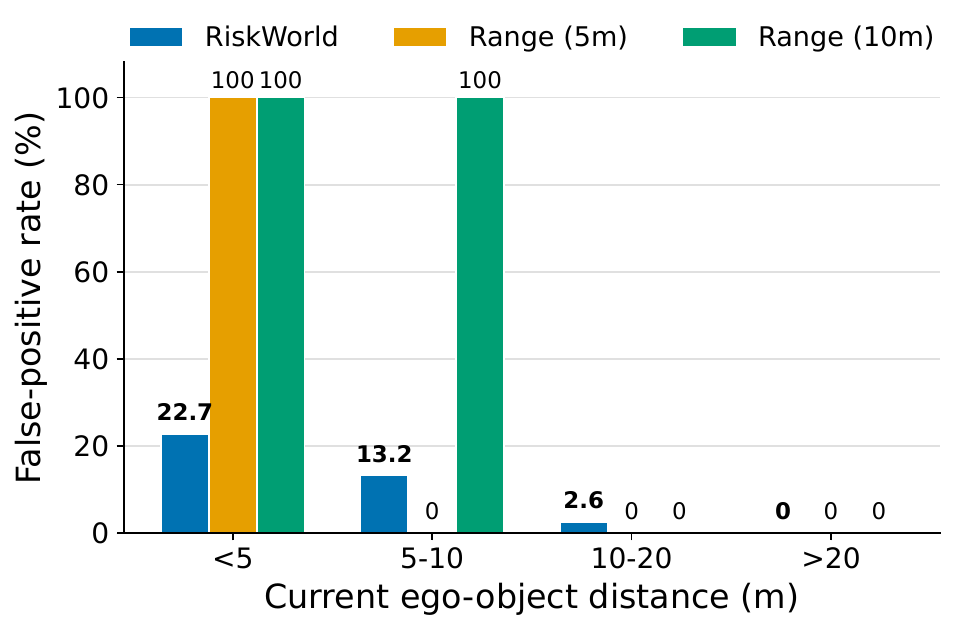}
\par\scriptsize (b) False-positive rate on non-risk objects
\end{minipage}\hfill
\begin{minipage}[t]{0.32\textwidth}
\vspace{0pt}
\centering
\includegraphics[width=\linewidth]{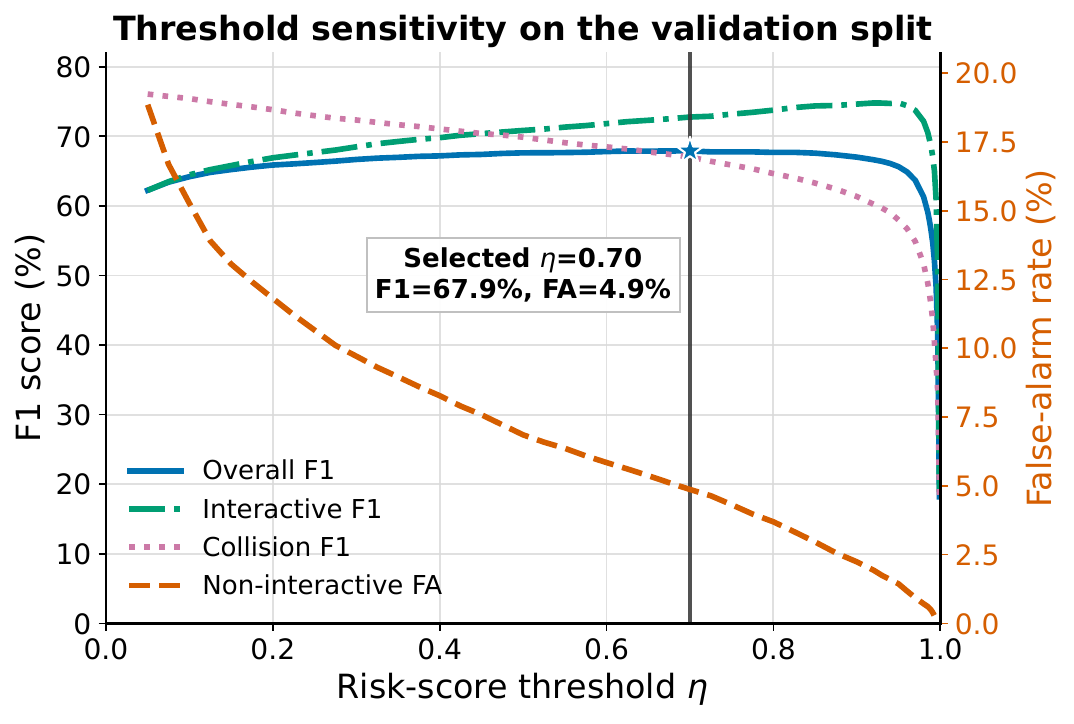}
\par\scriptsize (c) Operating-threshold sensitivity
\end{minipage}
\caption{\textbf{Risk-localization behavior and operating point.} (a–b) Test-set recall and false-positive rate across ego–object distances for RiskWorld and
fixed-range rules. (c) Overall, interactive, and collision F1 together with non-interactive false-alarm rate across validation thresholds; the marker denotes the selected $\eta=0.70$ used for all test categories.}
\label{fig:risk_localization_analysis}
\end{figure*}
\begin{figure}[t]
\centering
\begin{minipage}[t]{0.49\columnwidth}
\centering
\includegraphics[width=\linewidth]{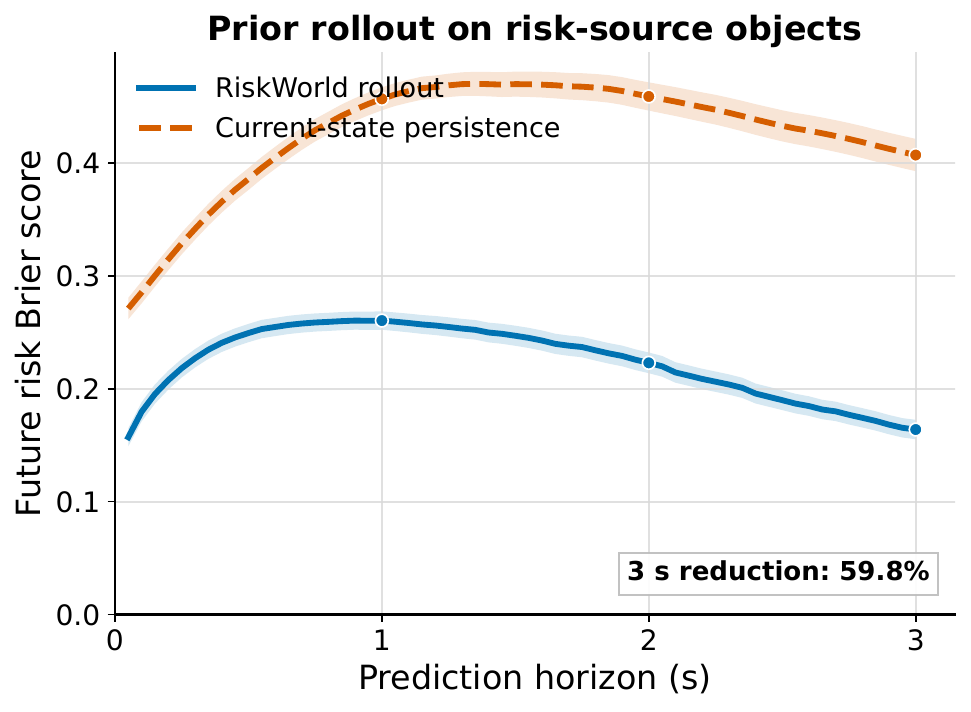}
\par\scriptsize (a) Future-rollout accuracy
\end{minipage}\hfill
\begin{minipage}[t]{0.49\columnwidth}
\centering
\includegraphics[width=\linewidth]{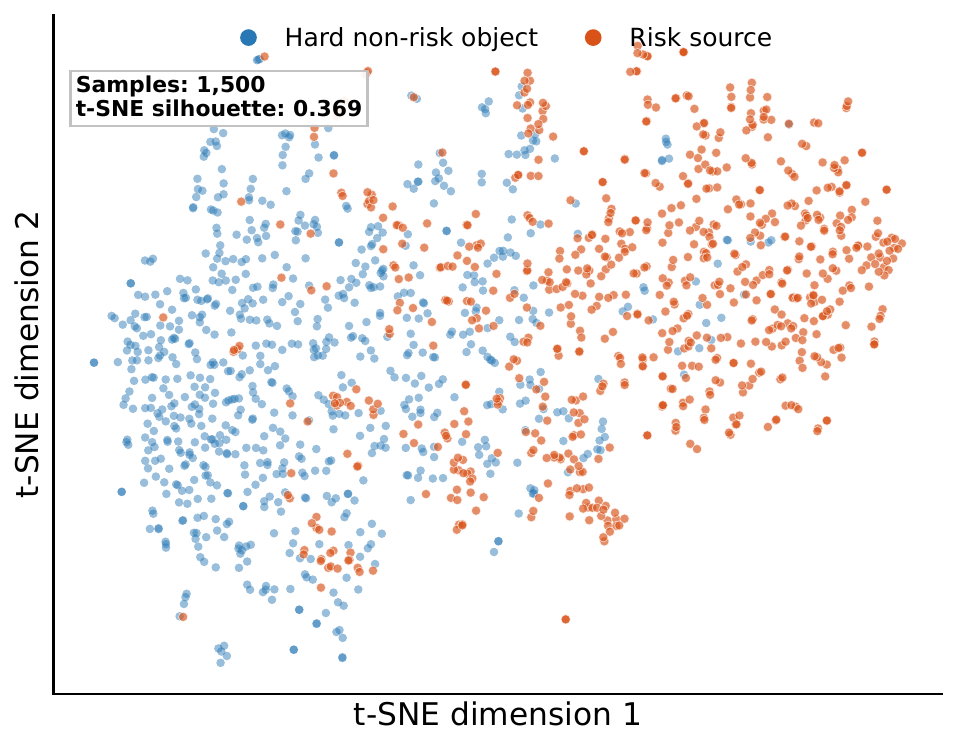}
\par\scriptsize (b) Decision-space separation
\end{minipage}
\caption{\textbf{Future-risk prediction and representation.} (a) Horizon-wise Brier score on risk-source objects, comparing the learned rollout with current-state persistence. Lower is better. (b) t-SNE of 1,500 balanced decision representations from risk sources and hard non-risk objects.}
\label{fig:future_risk_analysis}
\end{figure}

\subsection{Risk Localization and Anticipation}

To answer RQ1, we compare RiskWorld with representative rule-based, trajectory-based and behavior-based approaches.
Table~\ref{tab:riskbench_localization} shows that RiskWorld achieves the best overall F1 of 63.0, exceeding Range (10m), BADAS$^{*}$, and FLaRA$^{*}$ by 9.4, 14.2, and 1.2 points, respectively. FLaRA$^{*}$ is the strongest learned baseline, attaining 73.6\% recall on interactive scenes and an overall F1 of 61.8; this confirms that imagined future representations transfer effectively to risk localization. RiskWorld nevertheless achieves higher precision on all three risk-bearing subsets---70.2\% on interactive, 91.5\% on collision, and 67.3\% on obstacle---while attaining the lowest non-interactive FA of 2.1\%. Its advantage therefore lies in a stronger precision--recall--FA balance rather than uniformly higher recall. Range (10m) obtains higher recall and lower PIC on several subsets, but its broad proximity trigger increases FA to 15.2\% and reduces precision. RiskWorld instead offers the strongest overall F1 while more reliably distinguishing the risk source from irrelevant objects.

\textbf{Comparison with fixed-range rules.} Because Range (10m) remains competitive in recall and PIC, Figure~\ref{fig:risk_localization_analysis}(a--b) examines this comparison across ego--object distance intervals. A range rule selects every object within its radius and therefore incurs a 100\% false-positive rate in each active bin. RiskWorld recalls 76.7\% and 66.7\% of risk objects below 5\,m and between 5--10\,m while reducing the corresponding false-positive rates to 22.7\% and 13.2\%. It also retains 25.0\% recall at 10--20\,m, where both range rules are inactive. RiskWorld is therefore not acting as a learned range rule: it suppresses nearby distractors while retaining risk evidence beyond fixed distance cutoffs.

Figure~\ref{fig:risk_localization_analysis}(c) further examines sensitivity to the object-score threshold. Overall F1 remains stable over a broad interval around the selected $\eta=0.70$, while non-interactive FA decreases steadily as the threshold increases. This broad plateau shows that the reported performance does not depend on a narrowly tuned operating point and permits the recall--FA trade-off to be adjusted without retraining. 
% Together, the distance and threshold analyses show that RiskWorld remains selective across spatial ranges and robust to operating-point choice.

\subsection{Future-Risk Prediction Analysis}

To answer RQ2, we evaluate future-risk prediction against current-state persistence and examine whether risk sources separate from distractors before critical events.

\begin{figure}[t]
\centering
\includegraphics[width=\columnwidth]{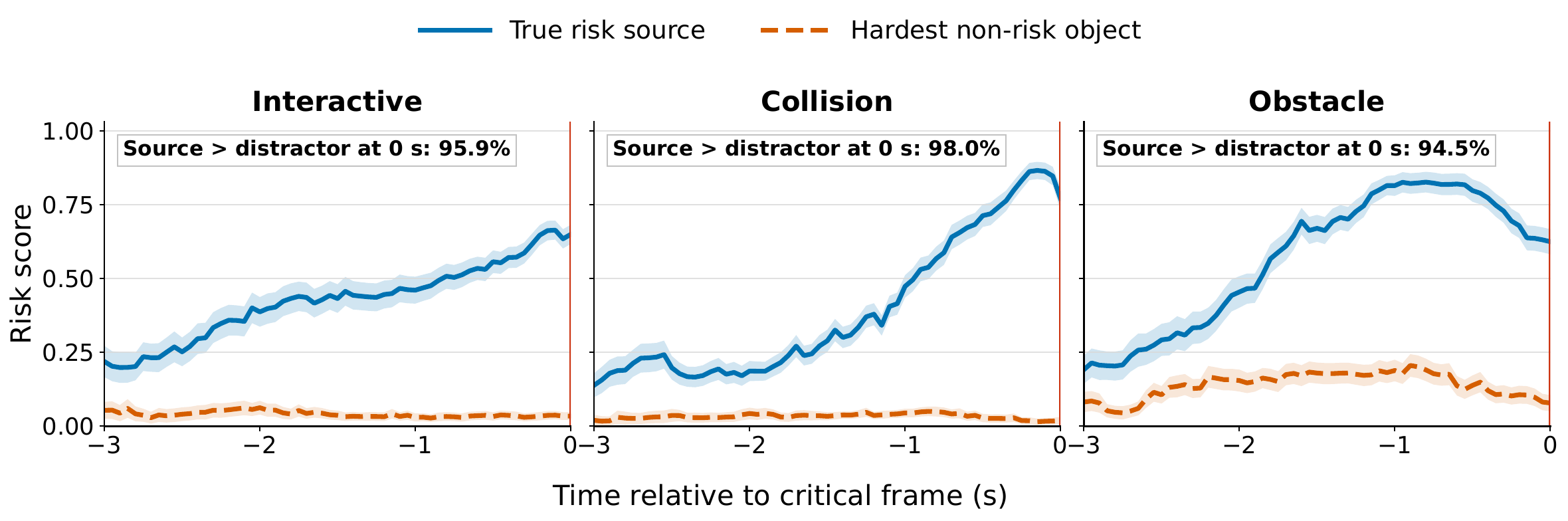}
\caption{\textbf{Temporal risk separation.} Mean scores of the annotated risk source and the strongest non-risk object from 3\,s before the critical frame; shaded regions denote 95\% confidence intervals.}
\label{fig:temporal_separation}
\end{figure}
\begin{table}[t]
\centering
\caption{Ablation of the main RiskWorld components. I, C, and O report F1 on interactive, collision, and obstacle scenarios; N reports the false-alarm rate on non-interactive scenes; All is overall micro-F1. All values are percentages.}
\label{tab:ablation}
\small
\begin{tabular}{lccccc}
\toprule
Variant & I $\uparrow$ & C $\uparrow$ & O $\uparrow$ & N $\downarrow$ & All $\uparrow$ \\
\midrule
w/o world features & 47.6 & 60.0 & 56.7 & 21.3 & 50.6 \\
w/o object visual token & 55.7 & 56.6 & 51.9 & 3.3 & 54.6 \\
w/o relation attention & 63.8 & 45.9 & 60.7 & 1.2 & 54.3 \\
Deterministic GRU & 65.3 & 63.9 & 68.5 & 6.0 & 62.0 \\
w/o future rollout & 62.4 & 59.9 & 64.4 & 5.9 & 60.2 \\
w/o future supervision & 68.5 & 57.5 & 65.4 & 4.4 & 61.6 \\
\midrule
RiskWorld & 68.3 & 56.0 & 69.5 & 2.1 & 63.0 \\
\bottomrule
\end{tabular}
\end{table}

Figure~\ref{fig:future_risk_analysis}(a) compares RiskWorld with a persistence baseline that repeats the current risk prediction at every future step. On annotated risk-source objects, RiskWorld achieves lower Brier scores across the full 3\,s horizon and reduces the error at 3\,s by 59.8\%, indicating that the rollout tracks future risk evolution more accurately than a static prediction.

Figure~\ref{fig:temporal_separation} further examines risk-source discrimination during the 3\,s preceding the critical frame. The mean source score rises from 0.219 to 0.650 on interactive scenes, from 0.139 to 0.771 on collision scenes, and from 0.191 to 0.625 on obstacle scenes, while the strongest non-risk score remains below 0.08 at the critical frame. At that frame, the annotated source outranks every distractor in 95.9\%, 98.0\%, and 94.5\% of the respective scenarios, showing increasingly clear separation as the event approaches. Finally, Figure~\ref{fig:future_risk_analysis}(b) shows distinct but partially overlapping decision regions; the two-dimensional embedding yields a silhouette score of 0.369 over 1,500 balanced samples. Together, these results show that latent rollout improves horizon-wise risk prediction and supports increasingly discriminative object-level decisions as critical events approach.

\begin{table}[t]
\centering
\caption{Planning-aware evaluation under the filtered-observation protocol. Lower values indicate that the selected object better preserves planning-critical information.}
\label{tab:planning_aware}
\resizebox{\columnwidth}{!}{
\begin{tabular}{lcccc}
\toprule
 & \multicolumn{2}{c}{Interactive} & \multicolumn{2}{c}{Obstacle} \\
\cmidrule(lr){2-3}\cmidrule(lr){4-5}
Method & IR $\downarrow$ & CR(\%) $\downarrow$ & IR $\downarrow$ & CR(\%) $\downarrow$ \\
\midrule
Auto-pilot & 0.45 & 33.6 & 0.52 & 58.7 \\
Full observation & 0.00 & 0.0 & 0.00 & 0.0 \\
Ground-truth risk & 0.02 & 0.4 & 0.04 & 5.3 \\
\midrule
Random & 0.11 & 7.7 & 0.44 & 48.1 \\
Range (10m) & \textbf{0.01} & 6.2 & 0.24 & 26.4 \\
Kalman filter & 0.06 & 6.9 & 0.33 & 41.8 \\
Social-GAN & 0.16 & 10.0 & 0.49 & 55.3 \\
MANTRA & 0.15 & 9.3 & 0.51 & 56.7 \\
QCNet & 0.05 & 10.0 & 0.52 & 57.2 \\
DSA & \textbf{0.01} & \textbf{0.8} & 0.37 & 38.5 \\
RRL & \textbf{0.01} & 1.2 & 0.49 & 54.3 \\
BP & 0.14 & 9.3 & 0.41 & 47.1 \\
BCP & 0.14 & 9.7 & 0.34 & 39.4 \\
\midrule
RiskWorld & 0.09 & 1.1 & \textbf{0.08} & \textbf{24.2} \\
\bottomrule
\end{tabular}
}
\end{table}
\begin{figure*}[t]
\centering
\begin{tabular}{ccc}
\includegraphics[width=0.32\textwidth]{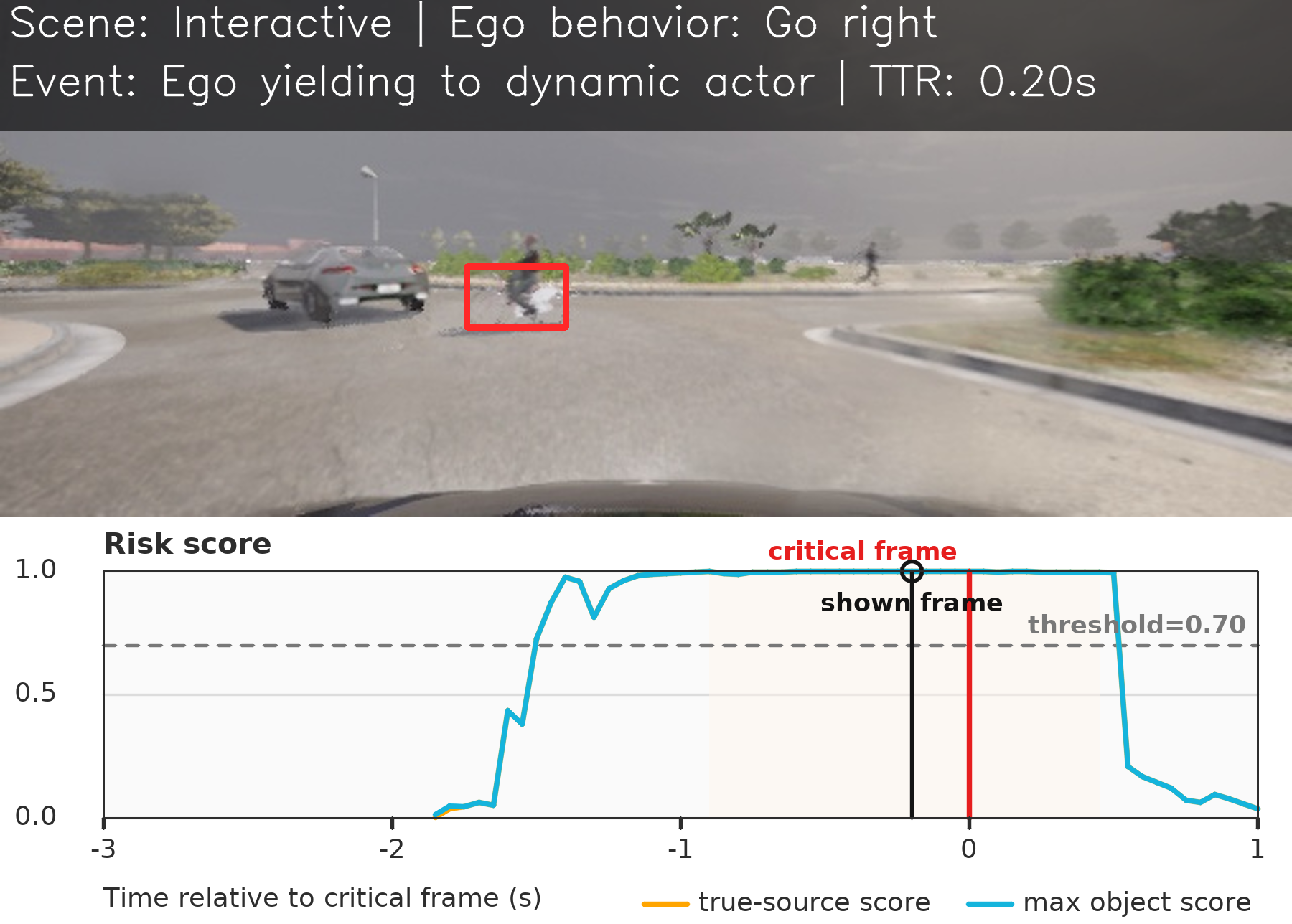} &
\includegraphics[width=0.32\textwidth]{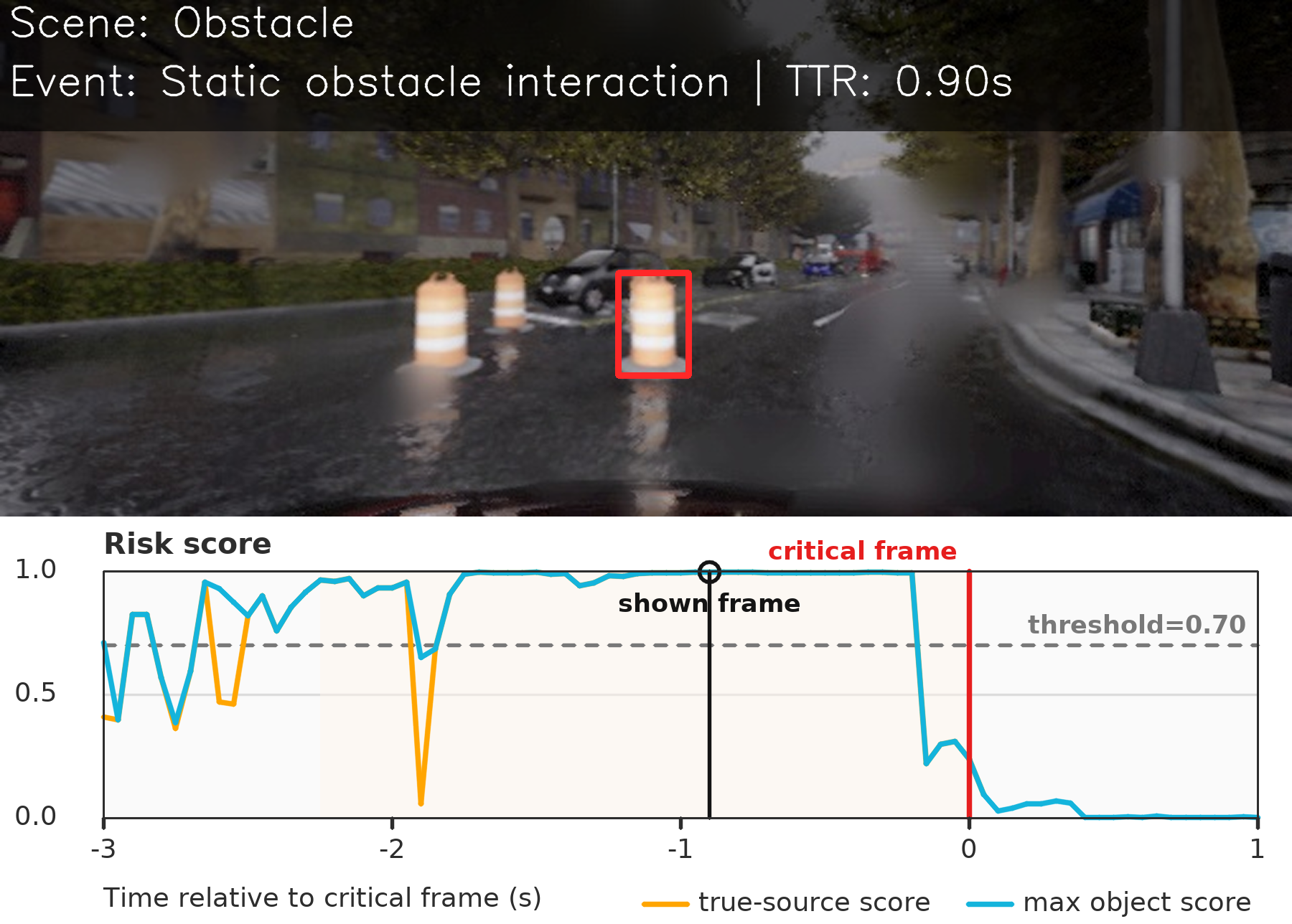} &
\includegraphics[width=0.32\textwidth]{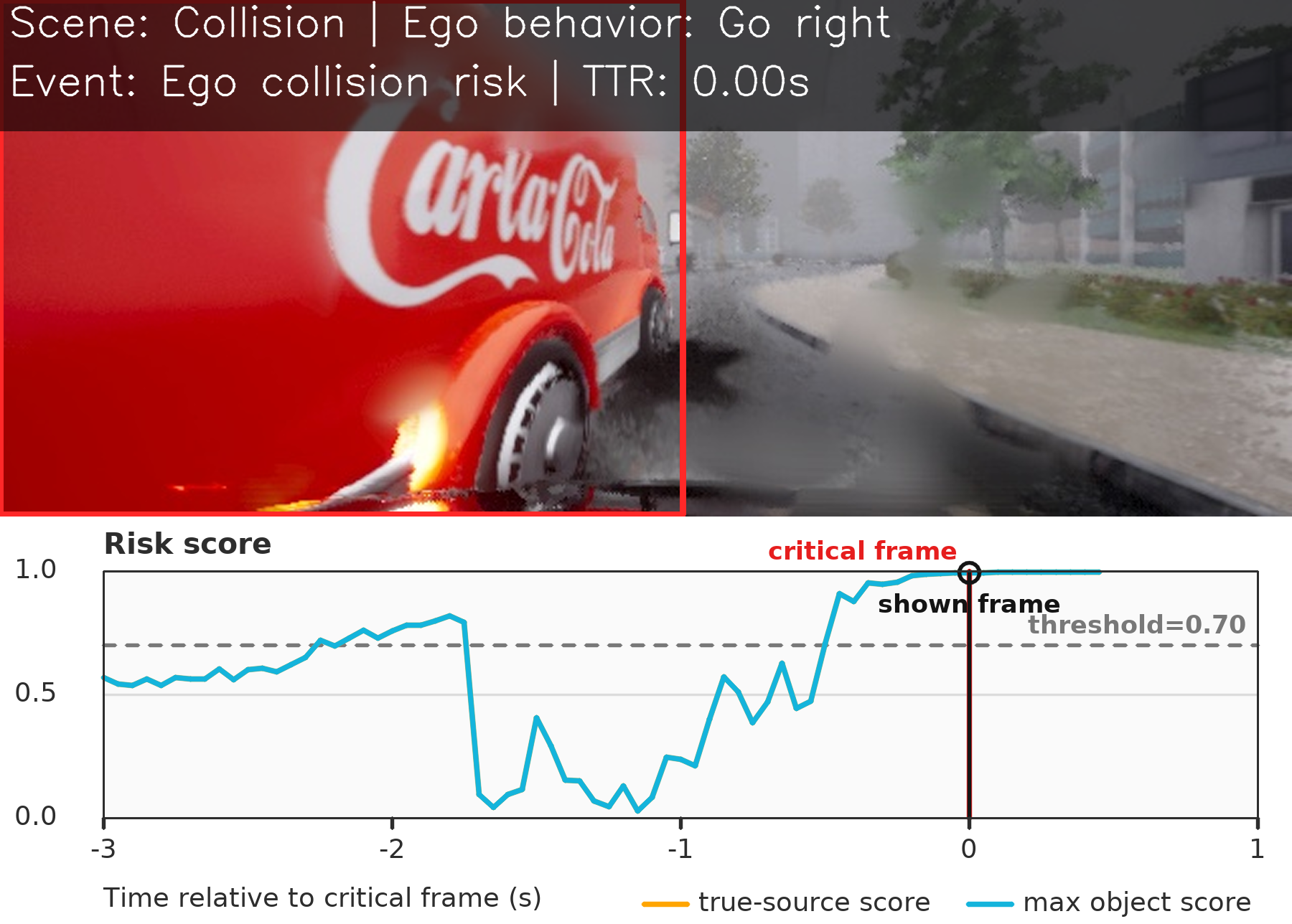} \\
(a) Interactive scenario & (b) Obstacle scenario & (c) Collision scenario
\end{tabular}
\caption{\textbf{Risk identification and temporal evolution.} The upper panels mark the predicted risk source in red. The lower panels show the annotated source score (orange) and maximum candidate score (cyan) from $-3$\,s to $+1$\,s relative to the critical frame. Black and red vertical lines indicate the displayed and critical frames, respectively. }
\label{fig:qualitative}
\end{figure*}

\subsection{Ablation Study}

To answer RQ3, Table~\ref{tab:ablation} compares six controlled variants in two groups. The representation group examines the contributions of pretrained scene context, object-aligned appearance, and relational reasoning. It removes all V-JEPA2 features (\emph{w/o world features}), removes bbox-aligned visual tokens while retaining the global scene feature (\emph{w/o object visual token}), or bypasses relational contextualization (\emph{w/o relation attention}). The future-modeling group separates the effects of stochastic dynamics, recursive latent rollout, and future-oriented supervision. It replaces the stochastic rollout with a parameter-matched recurrent model and removes $\mathcal{L}_{kl}$ (\emph{Deterministic GRU}), replaces recurrent rollout with a multi-horizon decoder while retaining future targets (\emph{w/o future rollout}), or removes $\mathcal{L}_{xy}$, $\mathcal{L}_{dist}$, $\mathcal{L}_{min}$, and $\mathcal{L}_{curve}$ while retaining $\mathcal{L}_{obj}$ and latent rollout (\emph{w/o future supervision}). All variants use the same experimental settings.
% , and each selects one validation threshold that is fixed across test categories.

\textbf{Object-centric representation.} The \emph{w/o world features} variant produces the largest degradation, reducing overall F1 from 63.0 to 50.6 and increasing FA from 2.1\% to 21.3\%. Retaining only the global scene feature recovers part of this loss, but \emph{w/o object visual token} remains 8.4 F1 points below RiskWorld, demonstrating the value of object-aligned visual grounding. The \emph{w/o relation attention} variant loses 8.7 F1 points; although its FA is lower, collision and obstacle F1 drop sharply. Thus, predictive scene context, object-specific appearance, and relational contextualization provide complementary evidence for localizing the risk source.

\textbf{Latent future modeling.} The \emph{Deterministic GRU} variant improves collision F1 but lowers overall F1 to 62.0 and raises FA to 6.0\%, indicating that stochastic dynamics provide a better aggregate trade-off. The \emph{w/o future rollout} variant further reduces overall F1 to 60.2 with 5.9\% FA, whereas \emph{w/o future supervision} yields 61.6 F1 with 4.4\% FA. Both recurrent imagination and future-oriented targets therefore contribute to object-level selectivity. Combining them with the object-centric representation gives RiskWorld the best overall F1 and a low FA of 2.1\%.

\subsection{Planning-Aware Evaluation}

% To answer RQ4, we follow the RiskBench filtered-observation protocol. The Learning by Cheating (LBC) planner \citep{lbc2020} receives only the objects selected by each method, while all other actors are masked. Full observation provides a no-bottleneck reference, whereas ground-truth risk provides a privileged selection reference. This experiment measures how much planning-relevant information the selections preserve; it neither evaluates RiskWorld as a planner nor claims improvements to LBC. 

To answer RQ4, an LBC planner \citep{lbc2020} observes either all objects, only the ground-truth risk object, or only the objects selected by each method; unselected actors are masked in the last setting. The first two provide full-observation and privileged-selection references. This comparison measures the planning relevance of the selected objects, not improvements to the planner itself.

As shown in Table~\ref{tab:planning_aware}, RiskWorld performs best among automatic selectors on obstacle scenarios, attaining an IR of 0.08 and a CR of 24.2\%. On interactive scenes, it achieves a low CR of 1.1\% but an IR of 0.09, indicating that unselected interaction context can still affect the planned trajectory. RiskWorld therefore preserves the most planning-relevant information among automatic selectors for obstacles, whereas interactive planning remains more dependent on broader scene context.

\subsection{Qualitative Analysis}

Figure~\ref{fig:qualitative} visualizes the frame-wise object risk scores produced by RiskWorld. At each observation time, the score of each candidate is decoded from its latent future rollout. In the interactive example, the crossing pedestrian's score rises sharply before the critical frame and remains near one as the interaction approaches. In the obstacle example, RiskWorld selects the relevant traffic barrel among nearby static objects and maintains high confidence through most of the pre-critical interval. The collision example is less stable at longer lead times, but its score increases rapidly as the collision approaches. Across the three scenarios, the source and maximum-candidate curves overlap over most high-confidence intervals, showing that the annotated source is generally the top-ranked candidate when RiskWorld reports high risk.

\section{Discussion and Limitations}

The results indicate that future ego--object relation rollout provides selective risk evidence beyond proximity, while the filtered-observation evaluation suggests that the selected objects retain planning-relevant information. RiskWorld nevertheless remains a risk monitor rather than a closed-loop planner; the planning-aware protocol evaluates the relevance of selected objects rather than direct improvements to planning. Current evaluation is limited to simulation; complex multi-object interactions, long-tail behaviors, and severe occlusions remain open challenges \citep{mmau2024}. Future work will study real-world generalization and uncertainty-aware, planner-conditioned counterfactual rollouts.

\section{Conclusion}

This paper presents RiskWorld, an object-centric latent world model for autonomous driving risk identification. RiskWorld constructs relation-aware object states from observed visual and motion histories, rolls them forward with RSSM-style latent dynamics, and decodes the imagined ego--object evolution into object-level risk scores and future-relation evidence. On RiskBench, it achieves 63.0\% overall F1 with a 2.1\% false-alarm rate; temporal and filtered-observation analyses further support early risk-source discrimination and the planning relevance of its selections. These results demonstrate the promise of object-centric latent world modeling for history-only driving risk monitoring.

\bibliography{aaai2027}

\end{document}